\documentclass{article}

\usepackage[preprint]{neurips_2023}

\usepackage[utf8]{inputenc} 
\usepackage[T1]{fontenc}    
\usepackage{url}            
\usepackage{booktabs}       
\usepackage{amsfonts}       
\usepackage{nicefrac}       
\usepackage{microtype}      
\usepackage{amsmath}
\usepackage{graphicx}
\usepackage{multirow}
\usepackage{float}
\usepackage{colortbl}
\usepackage{neurips_2023}
\usepackage{amssymb}

\usepackage{hyperref}
\hypersetup{colorlinks,allcolors=black}
\usepackage{tabulary}
\usepackage{makecell}
\usepackage{rotating}
\usepackage{caption}
\usepackage{subcaption}
\usepackage[table]{xcolor}
\usepackage{tikz}

\title{Temporal Contrastive Learning for Spiking Neural Networks}

\author{Haonan Qiu\textsuperscript{1}$^{*}$  \qquad Zeyin Song\textsuperscript{1}\thanks{: authors contributed equally. $^{\dag}$: corresponding author.}\qquad Yanqi Chen\textsuperscript{2,3} \qquad Munan Ning\textsuperscript{1,3} \qquad Wei Fang\textsuperscript{2,3}\\
 \textbf{Tao Sun\textsuperscript{1} \qquad Zhengyu Ma\textsuperscript{3}$^{\dag}$\qquad Li Yuan\textsuperscript{1,3}$^{\dag}$ \qquad Yonghong Tian\textsuperscript{1,2,3}$^{\dag}$}\\
\textsuperscript{1}School of Electronic and Computer Engineering, Peking University\\\
\textsuperscript{2}School of Computer Science, Peking University\\
\textsuperscript{3}Peng Cheng Laboratory\\
}

\begin{document}

\maketitle

\begin{abstract}

Biologically inspired spiking neural networks (SNNs) have garnered considerable attention due to their low-energy consumption and spatio-temporal information processing capabilities. Most existing SNNs training methods first integrate output information across time steps, then adopt the cross-entropy (CE) loss to supervise the prediction of the average representations. However, in this work, we find the method above is not ideal for the SNNs training as it omits the temporal dynamics of SNNs and degrades the performance quickly with the decrease of inference time steps. One tempting method to model temporal correlations is to apply the same label supervision at each time step and treat them identically. Although it can acquire relatively consistent performance across various time steps, it still faces challenges in obtaining SNNs with high performance. Inspired by these observations, we propose \textbf{T}emporal-domain supervised \textbf{C}ontrastive \textbf{L}earning (TCL) framework, a novel method to obtain SNNs with low latency and high performance by incorporating contrastive supervision with temporal domain information. Contrastive learning (CL) prompts the network to discern both consistency and variability in the representation space, enabling it to better learn discriminative and generalizable features. We extend this concept to the temporal domain of SNNs, allowing us to flexibly and fully leverage the correlation between representations at different time steps. Furthermore, we propose a \textbf{S}iamese \textbf{T}emporal-domain supervised \textbf{C}ontrastive \textbf{L}earning (STCL) framework to enhance the SNNs via augmentation, temporal and class constraints simultaneously. Extensive experimental results demonstrate that SNNs trained by our TCL and STCL can achieve both high performance and low latency, achieving state-of-the-art performance on a variety of datasets (\emph{e.g.}, CIFAR-10, CIFAR-100, and DVS-CIFAR10). 
\end{abstract}



\section{Introduction}


Over the past decade, artificial neural networks (ANNs) have achieved remarkable success in various computation vision tasks. However, with the rapid development of ANNs, the computational power they demand has increased significantly. Inspired by the brain's low energy and efficient computing capabilities, researchers have developed spiking neural networks (SNNs) with computational mechanisms inspired by the brain  to emulate these properties \cite{maass1997networks,gerstner2014neuronal,roy2019towards}. The brain-like properties of SNNs such as low energy consumption, event-driven nature, spatio-temporal information processing abilities, and inherent compatibility with neuromorphic hardware have drawn widespread attention. 

\begin{figure}[t]
\centering  
\includegraphics[scale=0.24]{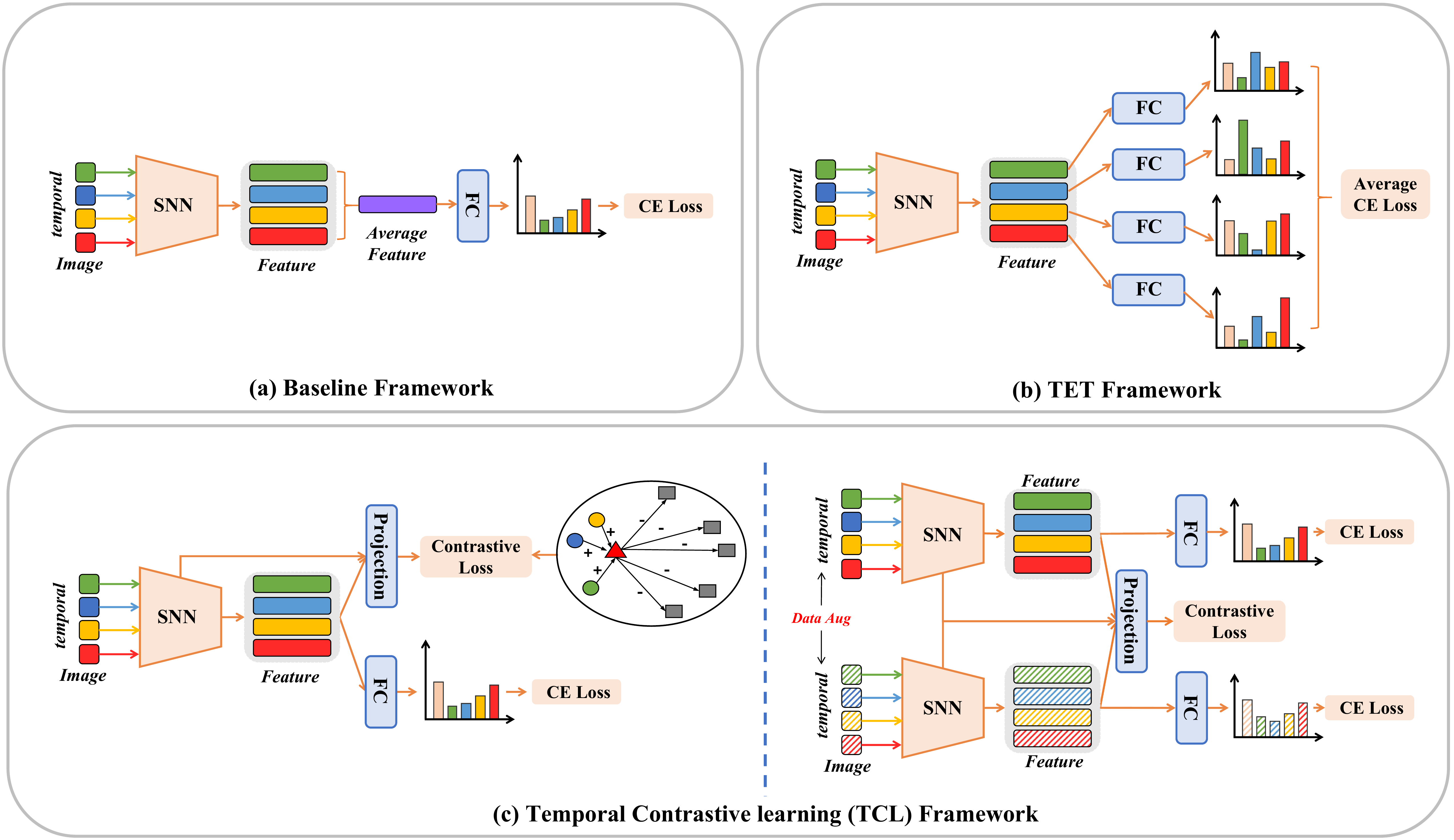}
\caption{An illustration of different methods for directly training SNNs. (a) Baseline framework: This method directly optimizes the aggregated representations under the label supervision. (b) TET framework: This approach optimizes the representations at each time step respectively, using identical label information. (c) Our proposed  \textbf{T}emporal-domain supervised \textbf{C}ontrastive \textbf{L}earning (TCL) framework (left) and \textbf{S}iamese \textbf{T}emporal-domain supervised \textbf{C}ontrastive \textbf{L}earning (STCL) framework (right): We inject contrastive supervision into temporal domain to model the correlation between representations at different time-steps.}
\vspace{-3mm}
\label{figure 1}
\end{figure}

\begin{figure}[ht]
  \centering
  \begin{subfigure}{0.40\textwidth}
    \centering
    \includegraphics[scale=0.40]{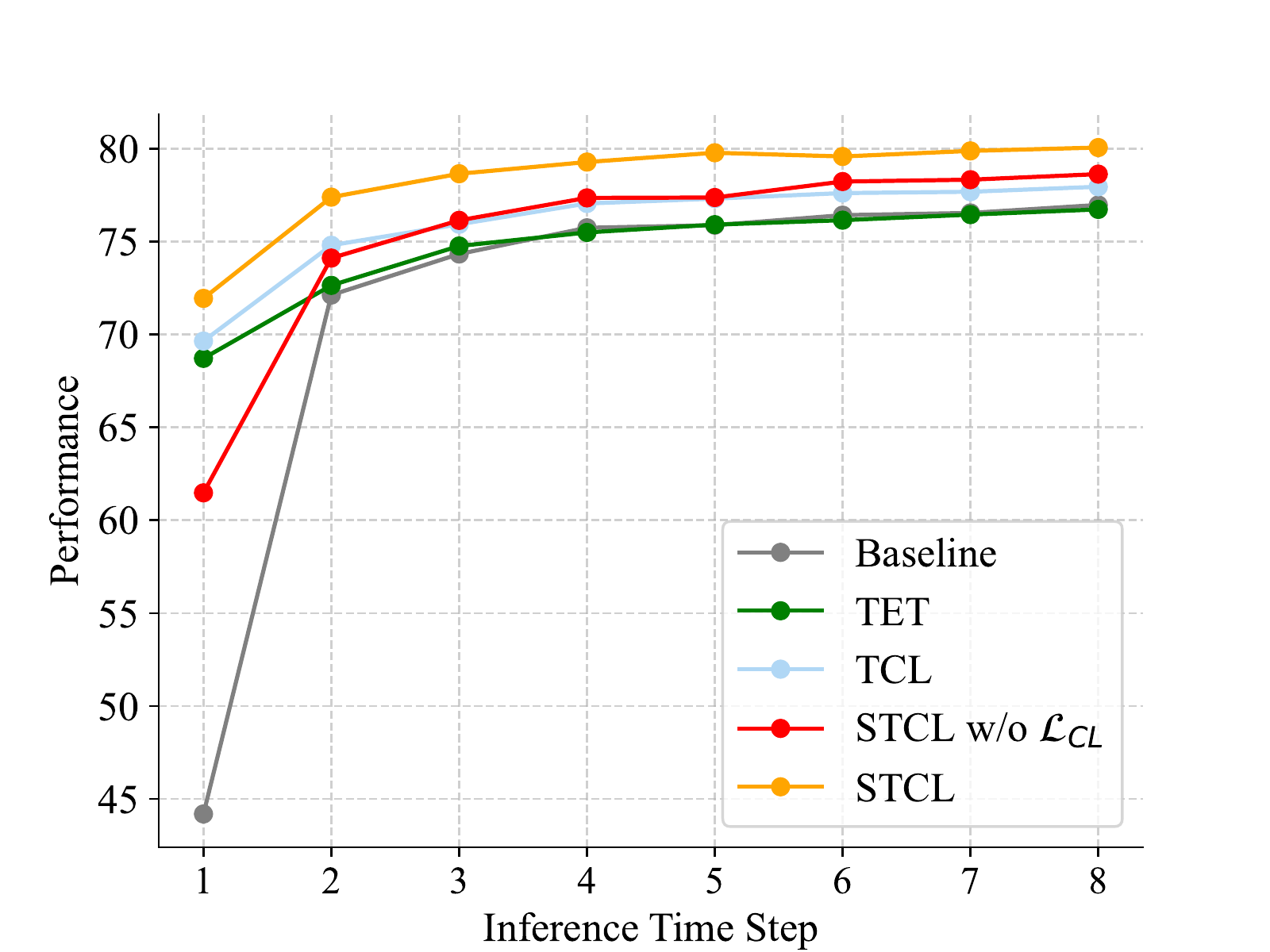}
    \caption{SEW-ResNet18.}
    \label{fig:tau}
  \end{subfigure}
   \hspace{0.02\textwidth}
  \begin{subfigure}{0.40\textwidth}
    \centering
    \includegraphics[scale=0.40]{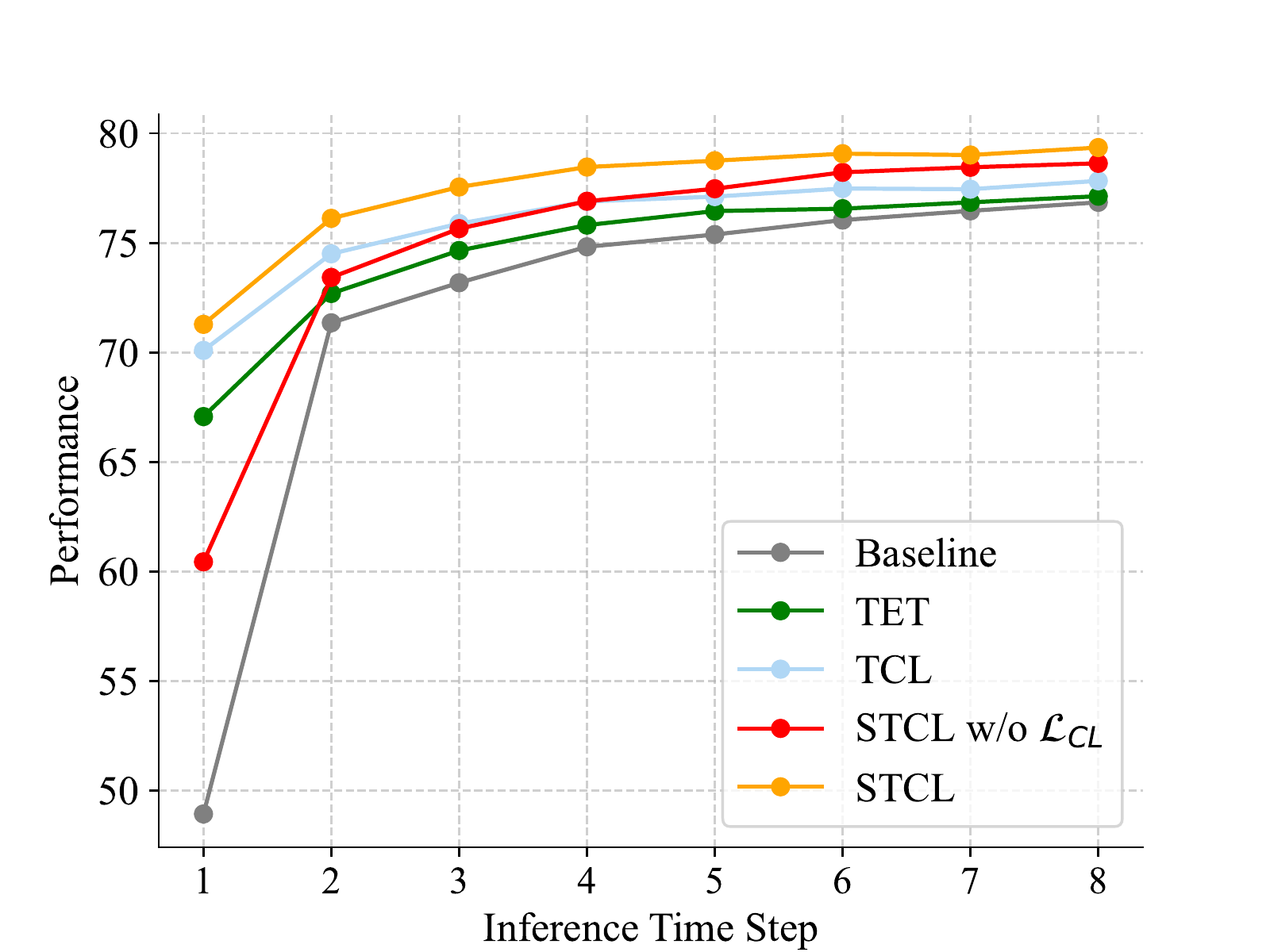}
    \caption{Spiking-ResNet18.}
    \label{fig:beta}
  \end{subfigure}
  \caption{Performance of SNNs at various time steps during the inference phase, with pre-training at time steps $T=8$ under CIFAR-100 dataset.}
  \label{fig:inference_time}
  \vspace{-5mm}
\end{figure}

Although SNNs have demonstrated notable potential, directly training high performance and low latency SNNs by backpropagation remains a challenge \cite{wu2018spatio,lee2016training,esser2015backpropagation,fang2021deep,mostafa2017supervised,neftci2019surrogate}. The mainstream approach involves using labeled information to supervise the average representations integrated over all time steps, defined as $\mathcal{L} = \mathcal{L}_{ce}(\frac{1}{T}\sum_{t=1}^{T}O^{t}, Y)$ (see Fig. \ref{figure 1}(a)). However, this method neglects the temporal dynamics of SNNs and only optimizes the average prediction distributions. As shown in Fig. \ref{fig:inference_time}, this strategy leads to a significant performance degradation when inference is conducted under ultra-low latency ($T=1$), indicating its inadequacy for scenarios with limited computing resources (\emph{i.e.}, 76.98\% -> 44.19\% for SEW \cite{fang2021deep} and 76.85\% -> 48.93\% for Spiking-ResNet \cite{hu2021spiking} in Fig. \ref{fig:inference_time}). 

A straightforward solution to make full use of all temporal information is to optimize each time step respectively, typically by assigning the same label supervision at each time step and treating them identically (\emph{i.e.}, TET \cite{deng2022temporal}), defined as $\mathcal{L} =\frac{1}{T}\sum_{t=1}^{T}\mathcal{L}_{ce}(O^{t}, Y)$ (see Fig. \ref{figure 1}(b)). Although it acquires relatively steady performance across various time steps (\emph{i.e.}, 76.98\% -> 68.72\% and 77.13\% -> 67.07\% in Fig. \ref{fig:inference_time}), this simplistic temporal supervision only takes temporal identity into account and overlooks coherent correlations. In addition, the self-accumulation intrinsic dynamics of spiking neurons and binary representation forms tend not to results in identical neural activities across time. All these factors limit the representational capability of SNNs and lead to minimal performance gains. The unsatisfactory performance of both the mainstream method and TET implies that the full power of temporal information remains to be harnessed through extra guidance remedy. 



To this end, we propose the temporal-domain supervised contrastive learning (TCL) framework to obtain SNNs with high performance and low latency by incorporating contrastive supervision signal with temporal domain information (see the left columns of Fig. \ref{figure 1}(c)). Contrastive learning (CL) is designed to learn both generalizable representations and discriminative representations \cite{Wu2018UnsupervisedFL,Ye2019UnsupervisedEL,Tian2019ContrastiveMC}. CL accomplishes this by attracting similar (positive) samples while dispelling different (negative) samples. In unsupervised contrastive learning, the positive samples could be another view of the anchor sample \cite{hadsell2006dimensionality,chen2020simple,zhang2022contrastive}, whereas in supervised contrastive learning, the positive samples are the samples sharing the same label with the anchor sample \cite{khosla2020supervised}. We extend this concept to the temporal domain of SNNs, constructing positives from features of 1) every single time step and 2) the sample within the same class. The SNNs trained by TCL flexibly and fully leverage the correlation between representations at different time steps, which
maintains high performance even under extremely low inference time steps (\emph{i.e.}, 77.96\% -> 69.65\% and 77.83\% -> 70.09\% in Fig. \ref{fig:inference_time}).


Furthermore, we propose a siamese temporal-domain supervised contrastive learning (STCL) framework (see the right columns of Fig. \ref{figure 1}(c)), to get more meaningful positive samples and facilitate the learning of extra data augmentation invariance. This extension encourages the SNNs to simultaneously learn data augmentation invariance, temporal correlation, and in-class correlation. Consequently, STCL effectively captures the spatio-temporal information of the SNNs, leading to superior and robust performance across different inference time steps (see Fig. \ref{fig:inference_time}).



Our experiments demonstrate the effectiveness of the proposed TCL and STCL frameworks, which achieve state-of-the-art performance on eight static and neuromorphic datasets. The main contributions of this work are as follows:

1) We first introduce contrastive learning to the representations in every single time step of SNNs and propose a temporal-domain supervised contrastive learning (TCL) framework to effectively supervise the temporal information of SNNs, enhancing their ability to model temporal correlation, overcoming the challenge of significant performance degradation during inference under ultra-low latency.

2) We further extend a siamese temporal-domain supervised contrastive learning (STCL) framework, which enhances the SNNs by learning invariance to data augmentation, the temporal and in-class correlation simultaneously. This significantly improves the performance of SNNs at low latency.

3) Extensive experiments on eight classification datasets, including CIFAR-10 \cite{krizhevsky2014cifar}, CIFAR-100, DVS-CIFAR10 \cite{li2017cifar10},  CUB-200-2011 \cite{wah2011caltech}, Stanford-Cars \cite{krause20133d}, FGVC-Aircraft \cite{maji2013fine}, Oxford Flowers \cite{nilsback2008automated}, and Stanford Dogs \cite{khosla2011novel}, demonstrate our proposed TCL and STCL outperform other approaches by a large margin and achieves the state-of-the-art (SOTA) performance.

\section{Related Work}

\subsection{Learning Methods of Spiking Neural Networks.}

The two primary training paradigms for SNNs are ANN-to-SNN conversion \cite{li2021free,deng2021optimal,ding2021optimal,han2020deep,han2020rmp} and Directly-train-SNN \cite{wu2018spatio,fang2021deep,shrestha2018slayer,meng2022training,zhou2022spikformer}. The conversion method transforms the pre-trained ANNs into the SNNs by replacing the ANN's activation function with spiking neurons and sharing weights. However, numerous time steps are required to match the spike firing frequency with the ANNs' activation values. Significant discrepancies can arise when the time step becomes exceedingly small, resulting in significant performance decline. Alternatively, the Directly-train-SNN strategy uses gradient backpropagation to train SNNs, creating a balance between performance and latency. This method supervises the temporal and spatial information of SNNs with backpropagation through time (BPTT) \cite{wu2018spatio,deng2022temporal,fang2021incorporating}, enabling the creation of low latency SNNs. Nonetheless, when compared with ANNs, there is still substantial potential for performance enhancement. This motivates us to propose more efficient methods for training low latency and high performance SNNs.

\subsection{Contrastive Learning.}

Contrastive learning has emerged as a compelling unsupervised paradigm, demonstrating its sophistication in representation learning \cite{chen2020simple,he2020momentum,Tian2019ContrastiveMC,He2019MomentumCF,Grill2020BootstrapYO,Caron2020UnsupervisedLO}. It designates positive and negative sample pairs in advance, narrows the distance between the representations of positive sample pairs using a contrastive loss function, and expands the distance between the representations of positive and negative sample pairs. Recently, researchers have incorporated label information into contrastive learning. This enhancement refines the definition of positive and negative samples, allowing contrastive learning to function within a supervised paradigm \cite{khosla2020supervised,zhang2022contrastive}, which further strengthens the model's capacity for representation. This study introduces this robust learning approach to effectively model the temporal correlations of SNNs, thereby elevating the representational capability of SNNs.


\section{Preliminary}


\subsection{Neuron Model}
In this work, we utilize Leaky Integrate-and-Fire (LIF) neuron as the foundational computational unit of the SNNs. The dynamic behavior of the neuron is described by the following equations:
\begin{equation}\label{eq1}
    u^{t+1, pre} = \alpha * u^{t} + x^{t+1},
\end{equation}
\begin{equation}\label{eq2}
    y^{t+1} = \Theta(u^{t+1, pre} - V_{th}),
\end{equation}
\begin{equation}\label{eq3}
    u^{t+1} = U^{t+1, pre} * (1-y^{t+1}),
\end{equation}
where $\alpha$ represents the delay factor (typically set to $0.5$ in our study), $u^{t+1, pre}$ signifies the membrane potential prior to the neuron's spike firing mechanism at time step $t+1$, and $u^{t}$ denotes the final membrane potential of the neuron at time step $t$. $x^{t+1}$ represents the input at time step $t+1$, while $\Theta$ denotes the step function, determining whether the neuron fires a spike based on whether the preliminary membrane potential surpasses the firing threshold $V_{th}$. If a spike is fired, the output $y^{t+1}$ is 1, otherwise, it is 0. Eq. \ref{eq3} illustrates the reset mechanism. Following a spike, the neuron's membrane potential is reset. In our model, we employ a 'hard reset' approach \cite{ledinauskas2020training}, resetting the membrane potential to a predetermined value, typically zero. This approach typically results in superior performance in the context of LIF neurons, compared to the 'soft reset' approach where a predetermined value (typically the firing threshold $V_{th}$) is subtracted from the membrane potential.

\section{Method}

In this section, we start by revisiting the loss function typically employed in the direct training of SNNs. Subsequently, we introduce our Temporal Contrastive Learning (TCL) framework, a novel approach aimed at effectively supervising the temporal information inherent in SNNs, thereby facilitating the network in modeling temporal correlation. Furthermore, we present our Siamese Temporal Contrastive Learning (STCL) framework, a siamese network paradigm to enhance SNNs to concurrently learn augmentation invariance, temporal and in-class correlation. We conclude this section by providing additional insights and strategies associated with our proposed frameworks.

\subsection{Direct Training of SNNs}


Direct training of SNNs predominantly uses the standard cross-entropy (CE) loss. A typical baseline method is first integrate output information across time steps, then adopt CE to supervise the prediction of the average representation (see Fig. \ref{figure 1}(a)):
\begin{equation}\label{eq:base_loss}
    \mathcal{L}_{{BL}} 
 = \mathcal{L}_{{CE}}\left(\frac{1}{T}\sum_{t=1}^{T}O^{t} , Y\right),
\end{equation}
where $O^{t}$ denotes the predicted probability output of the SNNs at time step $t$, $Y$ represents the target label, and $T$ is the total number of time steps. However, Eq. \ref{eq:base_loss} neglects the temporal dynamics of SNNs and only optimizes the temporal average prediction distributions. We demonstrate that SNNs trained by Eq. \ref{eq:base_loss} fails to maintain performance as time step decreasing during inference (see Fig. \ref{fig:inference_time}). 

Conversely, TET \cite{deng2022temporal} makes full use of all temporal information by assigning the same label supervision at each time steps and optimizing them respectively (see Fig. \ref{figure 1}(b)):
\begin{equation}\label{eq:tet_loss}
    \mathcal{L}_{{TET}} 
 = \frac{1}{T}\sum_{t=1}^{T} \mathcal{L}_{{CE}}(O^{t}, Y).
\end{equation}
Although Eq. \ref{eq:tet_loss} acquires relatively steady performance across various time steps (see Fig. \ref{fig:inference_time}), it indeed only takes temporal identity into account and overlooks coherent correlation, which leads to minimal performance gains and is also far from our objective, \emph{i.e.}, obtaining SNNs with both low latency and high performance.

\subsection{Temporal Contrastive Learning}





Contrastive learning is a powerful and flexible approach to representation learning. It learns to be discriminative and generalizable by maximizing the similarity of representations between positive samples and minimizing the similarity between positive and negative samples. By carefully defining positive and negative samples, and shaping the form of the contrastive loss function, we can equip the network with various capabilities. The common form of contrastive loss, such as SimCLR \cite{chen2020simple}, can be expressed as follows:

\begin{equation}\label{eq:contrastive_loss}
\mathcal{L}_{{CL}} = -\sum_{i \in I}\log\frac{\exp(z_{i} \cdot z_{i}^{+} / \tau )}{\sum_{k \neq i}\exp(z_{i} \cdot z_{k} / \tau)},
\end{equation}

where $I$ indexes all samples in a batch. The terms $z_{i}$, $z_{i}^{+}$, and $z_{k}$ all denote L2-normalized features that have been passed through a projection module. Similarly, all subsequent references to $z$ denote such features. Specifically, $z_{i}$ and $z_{i}^{+}$ represent the positive sample pairs, while $z_{k}$ represents the features of all other samples in the mini-batch that are not the $i$-th sample. $\tau$ is a temperature parameter that controls the concentration of the distribution. 

In our work, we incorporate contrastive supervision with temporal domain information and extend it to effectively capture the temporal correlations inherent in temporal information. We achieve this by redefining positive samples as representations of the same image at different time steps (see the left column of Fig. \ref{figure 1}(c)). This can be formalized as:


\begin{equation}\label{eq:CL_decomposed_corrected}
\mathcal{L}_{{CL}} = -\frac{1}{T}\sum_{{i \in I} }\sum_{t=1}^{T} \sum_{\substack{t=1\\t'\neq t}} \log \frac{\exp(z_{i}^{t} \cdot z_{i}^{t'} / \tau)}{\sum_{{(k,t'') \neq (i,t)}}\exp(z_{i}^{t} \cdot z_{k}^{t''} / \tau)},
\end{equation}

where $z_{i}^{t}$ denotes the representation of image $i$ at time step $t$, $T$ is the total number of time steps, and $t'$ denotes all other time steps for the same image $i$. The overall loss can be formulated as follow:



\begin{equation}\label{eq:total_loss}
    \mathcal{L}_{TCL} = \mathcal{L}_{{BL}} + \lambda \mathcal{L}_{{CL}},
\end{equation}

where the hyperparameter $\lambda$ balances the cross-entropy loss and the temporal contrastive loss.

\subsection{Siamese Network Paradigm}
In our previous discussions, we construct positive samples from features of every single time step. However, it has been established in prior research that the network can learn invariance to various forms of data augmentation, such as Color Jitter and Random Gray Scale \cite{howard2013some,chen2020simple,szegedy2015going}, thereby gaining a more generalized representational capacity. With this in mind, as shown in the right column of Fig. \ref{figure 1}(c), we adopt the siamese network paradigm \cite{chen2020simple,bromley1993signature} and simultaneously introduce data augmentation to further enhance the representational capacity of SNNs. We now define positive sample pairs to include different augmented versions of the same sample at various time steps. This updated approach guides the network to simultaneously capture temporal correlations across different time steps and maintain invariance to various data augmentations. This enhancement boosts its ability to seize and represent intricate temporal dynamics and transformations. The overall loss of the STCL framework can be formulated as follows:

\begin{equation}
\mathcal{L}_{{STCL}} = \mathcal{L}_{{BL}{aug1}} + \mathcal{L}_{{BL}{aug2}} + \lambda \mathcal{L}_{{CL}},
\end{equation}

where $\mathcal{L}_{{BL}{aug1}}$ and $\mathcal{L}_{{BL}{aug2}}$ represent the cross-entropy losses corresponding to the two different augmented versions of the same sample.

\subsection{Extra Details and Strategies}
{\bf Supervised Contrastive Loss.} In our supervised paradigm, we incorporate label information into the contrastive loss function, modifying it to include the similarity between samples of the same class, while still minimizing the similarity between positive and negative samples. Following the approach in \cite{khosla2020supervised}, the updated loss function can be formulated as:


\begin{equation} \label{supcon}
\mathcal{L}_{\ {CL}}^{\ {sup }} = -\frac{1}{T}\sum_{i \in I} \frac{1}{|P(i)|} \sum_{t=1}^{T} \sum_{t'=1}^{T} \sum_{p \in P(i)} \mathbb{1}{{i\neq p \text{ or } t\neq t'}} \log \frac{\exp(z_{i}^{t} \cdot z_{p}^{t'} / \tau)}{\sum_{{(k,t'') \neq (i,t)}}\exp(z_{i}^{t} \cdot z_{k}^{t''} / \tau)},
\end{equation}
where $P(i)$ denotes the set of positive sample pairs with the same label as sample $i$, and $|P(i)|$ represents the cardinality of the set $P(i)$. In this formulation, we extend the original contrastive loss function to include the similarity between samples of the same class at different time steps, while still minimizing the similarity between positive and negative samples. This adaptation allows the network to learn representations that more effectively capture the underlying structure of the temporal data, thereby leading to improved performance in classification tasks.

{\bf Projection Module.} In our framework, before calculating the temporal contrastive loss on the temporal features of SNNs, we introduce a projection module, which comprises two non-linear layers. This module maps the features of the SNNs into a normalized embedding space. This pivotal component significantly enhances the model's performance, corroborating findings in contrastive learning \cite{chen2020simple} and knowledge distillation \cite{chen2022improved}. Throughout the training process, we incorporate the projection module, but it is omitted during inference, thereby adding no additional parameters. 

{\bf Deep Contrastive Supervision.} Additionally, inspired by the work presented in \cite{zhang2022contrastive}, beyond supervising the temporal information in the last layer of the SNNs, we also apply contrastive loss calculations to the intermediate layers of the SNNs by default. This strategy provides effective supervisory signals in the intermediate layers for training the SNNs.


\section{Experiments}



In this section, we present a variety of experiments to validate the effectiveness of our proposed method. The experiments are designed to cover multiple datasets, including three static and neuromorphic datasets (\emph{e.g.}, CIFAR-10, CIFAR-100, and DVS-CIFAR10) and five fine-grained datasets (\emph{e.g.}, CUB-200-2011(CUB), Stanford-Cars(Car), FGVC-Aircraft(Aircraft), Oxford Flowers(Flowers), and Stanford Dogs(Dogs)), and leverage widely used base models, Spiking-ResNet \cite{zheng2021going}, SEW-ResNet\cite{fang2021deep} and Spiking-VGG \cite{deng2022temporal}. We first describe the dataset preprocessing methods and implementation details to provide a clear understanding of the experimental setup. We then delve into the application of our proposed method on mainstream static and neuromorphic image classification tasks to demonstrate its performance in the SNNs standard tasks scenario. Following this, we explore the fine-grained image classification capabilities of our proposed method. Further, we compare our method with the other SOTA training methods. Finally, we conduct a series of ablation studies to gain deeper insights into the contributions of various components of our proposed method.

\subsection{Dataset Preprocessing and Implementation Details}

In line with previous setups, we directly feed the original images into the SNN for $T$ times on static and fine-grained datasets, while on DVS datasets, we directly input the original spike data. By default, we set the time step $T$ for SNN training and inference to 4.

{\bf Data Augmentation}. 
Regarding the static and fine-grained datasets, we employ standard data augmentation techniques for the input images, including random cropping, flipping, and normalization in the TCL framework. To enhance our STCL framework, we introduce more complex data augmentation techniques such as Cutout \cite{devries2017improved} \emph{(only for CIFAR-10/100)}, Color Jitter, and Random Gray Scale. For neuromorphic datasets DVS-CIFAR10, we reduce the resolution of the input image from 128 $\times$ 128 to 48 $\times$ 48 and apply random cropping as the augmentation strategy in our frameworks.

{\bf Training}. We train our models for 200 epochs using SGD optimizer with a learning rate of 0.1, a weight decay of 0.0005, and the momentum of 0.9, using a batch size of 64 or 128, on all datasets.


\subsection{Mainstream Static and Neuromorphic Image Classification}

In this subsection, we evaluate the effectiveness of our proposed method, including both the TCL and STCL frameworks, on mainstream static image classification tasks (CIFAR-100 and CIFAR-10) and a neuromorphic dataset (DVS-CIFAR10). The results are shown in Tab. \ref{Table Effectiveness of our proposed TCL and STCL framework.}, where we compare the performance of our method using SEW-ResNet18 (SEW-18) and SEW-ResNet34 (SEW-34) architectures against the baseline performance. The results indicate that our proposed TCL and STCL methods consistently improve classification accuracy across all datasets and architectures. Specifically, for the CIFAR-100 dataset, the STCL achieves the highest accuracy of 79.51\% and 80.30\% for SEW-18 and SEW-34, respectively, marking an improvement of 3.44\% and 3.16\% over the baseline. Moreover, on the CIFAR-10 dataset, the STCL significant outperforms the baseline by achieving 95.97\% and 96.39\% accuracy for SEW-18 and SEW-34, respectively. Looking at the DVS-CIFAR10 dataset, the TCL lifts the classification accuracy from 75.60\% to 77.20\% for SEW-18. For the SEW-34, the TCL increases the accuracy from 74.70\% to 76.60\%, and the STCL achieves an even higher accuracy of 78.30\%.

Absolutely, the STCL without the contrastive loss ($\mathcal{L}_{\text{CL}}$ in Eq. \ref{supcon}), denoted as STCL w/o $\mathcal{L}_{\text{CL}}$, achieves 78.60\% and 79.63\% accuracy for SEW-18 and SEW-34 on CIFAR-100, and 95.44\% and 96.07\% on CIFAR-10, respectively. This implies that, even in the absence of the contrastive loss, the network can benefit from learning to be invariant to data augmentations such as Random Gray Scale. However, by employing the contrastive loss, our full STCL can bring about further improvements by learning both augmentation invariance and temporal correlation. These results demonstrate that our framework, through effectively learning augmentation invariance and temporal correlation, consistently brings about stable performance improvements across various datasets. 

\vspace{-2mm}

\begin{table}[ht]
\centering
\caption{Effectiveness of our proposed TCL and STCL framework on CIFAR and DVS-CIFAR.}
\label{Table Effectiveness of our proposed TCL and STCL framework.}
\setlength\tabcolsep{1.9mm}
\renewcommand\arraystretch{1.1}
{
\scalebox{0.85}{
\begin{tabular}{ccccccc}
\toprule
Dataset & \multicolumn{2}{c}{CIFAR-100} & \multicolumn{2}{c}{CIFAR-10} &\multicolumn{2}{c}{DVS-CIFAR10}\\ \midrule
Model        & SEW-18        & SEW-34        & SEW-18        & SEW-34    & SEW-18        & SEW-34    \\ \midrule
\rowcolor{gray!20} Baseline     & 76.07         & 77.14         &   94.47            &   94.48     &   75.60           &   74.70      \\ \midrule
TCL          & {77.38}(\textcolor{red}{+1.31})    & {77.84}(\textcolor{red}{+0.70})   &    {94.92}(\textcolor{red}{+0.45})    &   {94.95}(\textcolor{red}{+0.47})   &    \textbf{77.20}(\textcolor{red}{+1.60})   &   {76.60}(\textcolor{red}{+1.90})      \\ \midrule
STCL w/o $\mathcal{L}_{\text{CL}}$    & 78.60(\textcolor{red}{+2.53})  & 79.63(\textcolor{red}{+2.49})   & {95.44}(\textcolor{red}{+0.97})    &  {96.07}(\textcolor{red}{+1.59})   &    \emph{74.70}(\textcolor{green}{-0.90})   &   {75.50}(\textcolor{red}{+0.80})      \\ \midrule
STCL & \textbf{79.51}(\textcolor{red}{+3.44})  & \textbf{80.30}(\textcolor{red}{+3.16})   & \textbf{95.97}(\textcolor{red}{+1.5})   &   \textbf{96.39}(\textcolor{red}{+1.91})   &    {76.60}(\textcolor{red}{+1.00})    &   \textbf{78.30}(\textcolor{red}{+3.60})
\\ \bottomrule
\end{tabular}
}}
\vspace{-2mm}
\end{table}

\subsection{Fine-Grained Image Classification}


In this section, we assess the performance of our proposed frameworks on five fine-grained image classification tasks, utilizing SEW-ResNet18 as our base model. Tab. \ref{Fine-Gained Image Classfication} showcases the classification accuracy for each method. The STCL framework consistently surpasses the baseline across all datasets. Specifically, it shows improvements of 6.98\% on CUB (57.46\% vs. 50.48\%), 5.42\% on Car (83.37\% vs. 77.95\%), and 5.85\% on Aircraft (70.15\% vs. 64.30\%). Meanwhile, the TCL framework also demonstrates robust performance enhancements on four datasets. The outcomes affirm the effectiveness of our methods in leveraging augmentation invariance and temporal correlation under fine-grained datasets, thereby enhancing the representational capacity of SNNs. 
\vspace{-2mm}



\begin{table}[htbp]
\centering
\caption{Fine-Gained Image Classfication. Numbers following datasets indicate category count.}
\label{Fine-Gained Image Classfication}
\setlength\tabcolsep{1.9mm}
\renewcommand\arraystretch{1.1}
{
\scalebox{0.85}{
\begin{tabular}{cccccc}
\toprule
Dataset  & CUB$_{200}$ & Car$_{196}$ & Aircraft$_{102}$ & Flowers$_{102}$ & Dogs$_{120}$ \\ \midrule
\rowcolor{gray!20} BaseLine & 50.48 & 77.95    &  64.30  & 89.90  & 58.79      \\ \midrule
TCL      & \emph{50.29}(\textcolor{green}{-0.19})    & 78.73(\textcolor{red}{+0.78})    &  66.37(\textcolor{red}{+2.07}) & 90.00(\textcolor{red}{+0.10}) & 59.16(\textcolor{red}{+0.37}) \\ \midrule
STCL     &  \textbf{57.46}(\textcolor{red}{+6.98})   & \textbf{83.37}(\textcolor{red}{+5.42})     &  \textbf{70.15}(\textcolor{red}{+5.85})         &   \textbf{94.90}(\textcolor{red}{+5.00})       &  \textbf{61.24}(\textcolor{red}{+2.45})   \\
\bottomrule
\end{tabular}
}}
\vspace{-2mm}
\end{table}

\subsection{Comparison with state-of-the-art SNN Training Methods}
In this section, we compare our proposed methods with current state-of-the-art SNN training methods on CIFAR-10, CIFAR-100, and DVS-CIFAR10 datasets.  The results are presented in Tab. \ref{Comparison with the other method on Datasets}.

\begin{table}[ht]
\centering
\caption{Comparisons with current state-of-the-art methods on CIFAR and DVS-CIFAR.}
\label{Comparison with the other method on Datasets}
\setlength\tabcolsep{1.9mm}
\renewcommand\arraystretch{1.1}
{
\scalebox{0.95}{
\begin{tabular}{cccccc}
\toprule
Dataset     & {Method}   & Training-SNN    & Model        & Time step & Accuracy \\
\midrule

\multirow{7}[2]{*}{\emph{CIFAR-10}}

& TET\cite{deng2022temporal}\textsuperscript{\textcolor{gray}{ICLR2022}} & \textbf{\text{\sffamily\bfseries\Large\checkmark}}    & ResNet-19     & 6         & 94.50     \\
& Rec-Dis\cite{guo2022recdis}\textsuperscript{\textcolor{gray}{CVPR2022}}  & \textbf{\text{\sffamily\bfseries\Large\checkmark}}      & ResNet-19     & 6         & 95.55    \\

& TEBN\cite{duan2022temporal}\textsuperscript{\textcolor{gray}{NeurIPS2022}} & \textbf{\text{\sffamily\bfseries\Large\checkmark}}     & ResNet-19     & 6         & 95.60   \\ 
& GLIF\cite{yao2022glif}\textsuperscript{\textcolor{gray}{NeurIPS2022}} &\textbf{\text{\sffamily\bfseries\Large\checkmark}}     & ResNet-19     & 6         & 95.03   \\ 
& ESG\cite{guo2022loss}\textsuperscript{\textcolor{gray}{NeurIPS2022}}  & \textbf{\text{\sffamily\bfseries\Large\checkmark}}    & ResNet-19     & 6         & 95.49   \\ 
\cmidrule{2-6}

& {TCL}  &\multirow{2}{*}{\textbf{\text{\sffamily\bfseries\Large\checkmark}}}  & \multirow{2}{*}{{ResNet-19}}     & \multirow{2}{*}{\textbf{4}}         & \textbf{95.03}  \\ 

&{STCL}  &  &  &                & \textbf{96.35}  \\ 
\midrule
\multirow{6}[2]{*}{\emph{CIFAR-100}} 
&  TET\cite{deng2022temporal}\textsuperscript{\textcolor{gray}{ICLR2022}} & \textbf{\text{\sffamily\bfseries\Large\checkmark}}    & ResNet-19     & 6         & 74.72    \\
&  Rec-Dis \cite{guo2022recdis}\textsuperscript{\textcolor{gray}{CVPR2022}}   & \textbf{\text{\sffamily\bfseries\Large\checkmark}}      & ResNet-19     & 6         & 74.10     \\
& TEBN\cite{duan2022temporal}\textsuperscript{\textcolor{gray}{NeurIPS2022}} & \textbf{\text{\sffamily\bfseries\Large\checkmark}}     & ResNet-19     & 6         & 78.76  \\ 
& GLIF\cite{yao2022glif}\textsuperscript{\textcolor{gray}{NeurIPS2022}}  & \textbf{\text{\sffamily\bfseries\Large\checkmark}}    & ResNet-19     & 6         & 77.35   \\ 
\cmidrule{2-6}
& {TCL}   &\multirow{2}{*}{\textbf{\text{\sffamily\bfseries\Large\checkmark}}}  & \multirow{2}{*}{{ResNet-19}}     & \multirow{2}{*}{\textbf{4}}         & \textbf{79.73}  \\ 
& {STCL}  &  &   &               & \textbf{81.84}  \\ \midrule

\multirow{8}[3]{*}{\emph{DVS-CIFAR10}} 
& Dspike\cite{li2021differentiable}\textsuperscript{\textcolor{gray}{NeurIPS2021}}     & \textbf{\text{\sffamily\bfseries\Large\checkmark}}    & ResNet18     & 10         & 75.40    \\
&  TET\cite{deng2022temporal}\textsuperscript{\textcolor{gray}{ICLR2022}}      &\textbf{\text{\sffamily\bfseries\Large\checkmark}}    & VGG   & 10         & 83.17    \\
&  Rec-Dis \cite{guo2022recdis}\textsuperscript{\textcolor{gray}{CVPR2022}}      & \textbf{\text{\sffamily\bfseries\Large\checkmark}}     & ResNet-19     & 10         & 72.42     \\
&  DSR\cite{meng2022training}\textsuperscript{\textcolor{gray}{CVPR2022}}     &\textbf{\text{\sffamily\bfseries\Large\checkmark}}    & VGG & 20        & 77.27     \\
& TEBN\cite{duan2022temporal}\textsuperscript{\textcolor{gray}{NeurIPS2022}}    & \textbf{\text{\sffamily\bfseries\Large\checkmark}}    & VGG     & 10         & 84.90  \\ 
\cmidrule{2-6}
& {TCL}  & \multirow{2}{*}{\textbf{\text{\sffamily\bfseries\Large\checkmark}}}   & \multirow{2}{*}{{VGG}}     & \multirow{2}{*}{\textbf{4}}         & \textbf{79.10}  \\ 
& {STCL}       &  &     &        & \textbf{79.80}  \\ 

\bottomrule
\end{tabular}
}}
\vspace{-2mm}
\end{table}

\noindent {\bf CIFAR-10}. Our STCL method achieves an accuracy of 96.35\%, which is the highest among all the compared methods. It outperforms the second-best method, Rec-Dis \cite{guo2022recdis}, by 0.8\%. Furthermore, our TCL method also achieves competitive performance with an accuracy of 95.03\%, which is on par with GLIF \cite{yao2022glif} and ESG \cite{guo2022loss}. Notably, our methods utilize only 4 time steps, which is lower than most of the other methods that use 6 time steps. This reduction in time steps indicates the efficiency of our proposed methods.

\noindent {\bf CIFAR-100.} Our STCL and TCL method ranks higher than all of the compared methods and STCL achieves the best accuracy of 81.84\%, outperforming the second-best method, TEBN \cite{duan2022temporal}, by 3.08\%. Again, our methods require only 4 time steps, highlighting their efficiency in SNN training.

\noindent {\bf DVS-CIFAR10.} On the DVS-CIFAR10 dataset, our TCL method achieves an accuracy of 79.10\%, while our STCL method reaches an accuracy of 79.80\%. Compared to the other methods, our methods perform competitively. Our STCL method surpasses the performance of DSR, but falls short of TEBN. However, it is worth noting that our methods only require 4 time steps, which is significantly fewer than most of the other methods that use 10 or more time steps. Our TCL method outperforms Dspike \cite{li2021differentiable}, Rec-Dis, and DSR, which highlights the effectiveness of our proposed framework.



\subsection{Ablation Studies}

\begin{figure}[ht]
  \centering
  \begin{subfigure}{0.45\textwidth}
    \centering
    \includegraphics[scale=0.40]{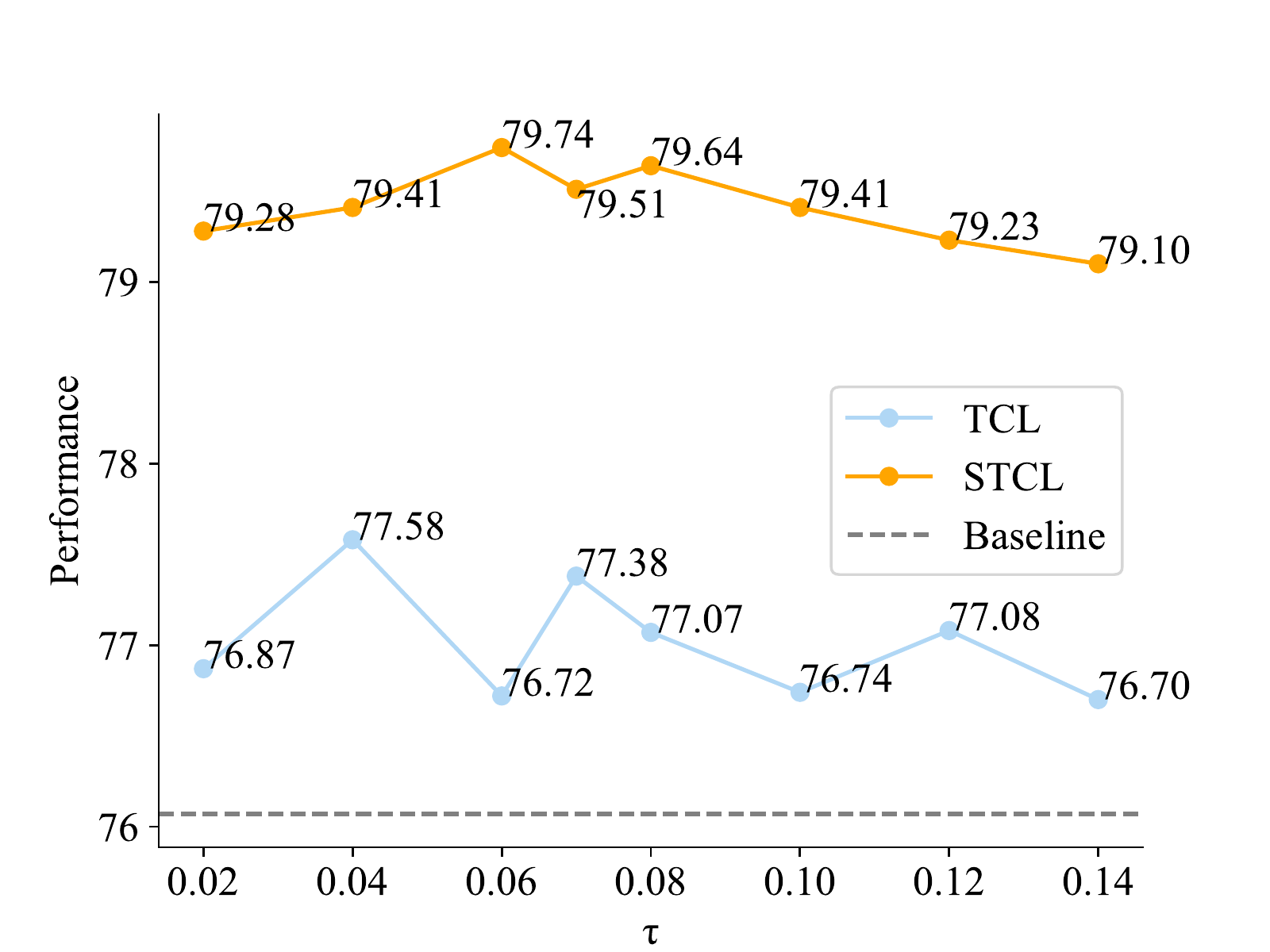}
  \end{subfigure}
  \hspace{0.02\textwidth}
  \begin{subfigure}{0.45\textwidth}
    \centering
    \includegraphics[scale=0.40]{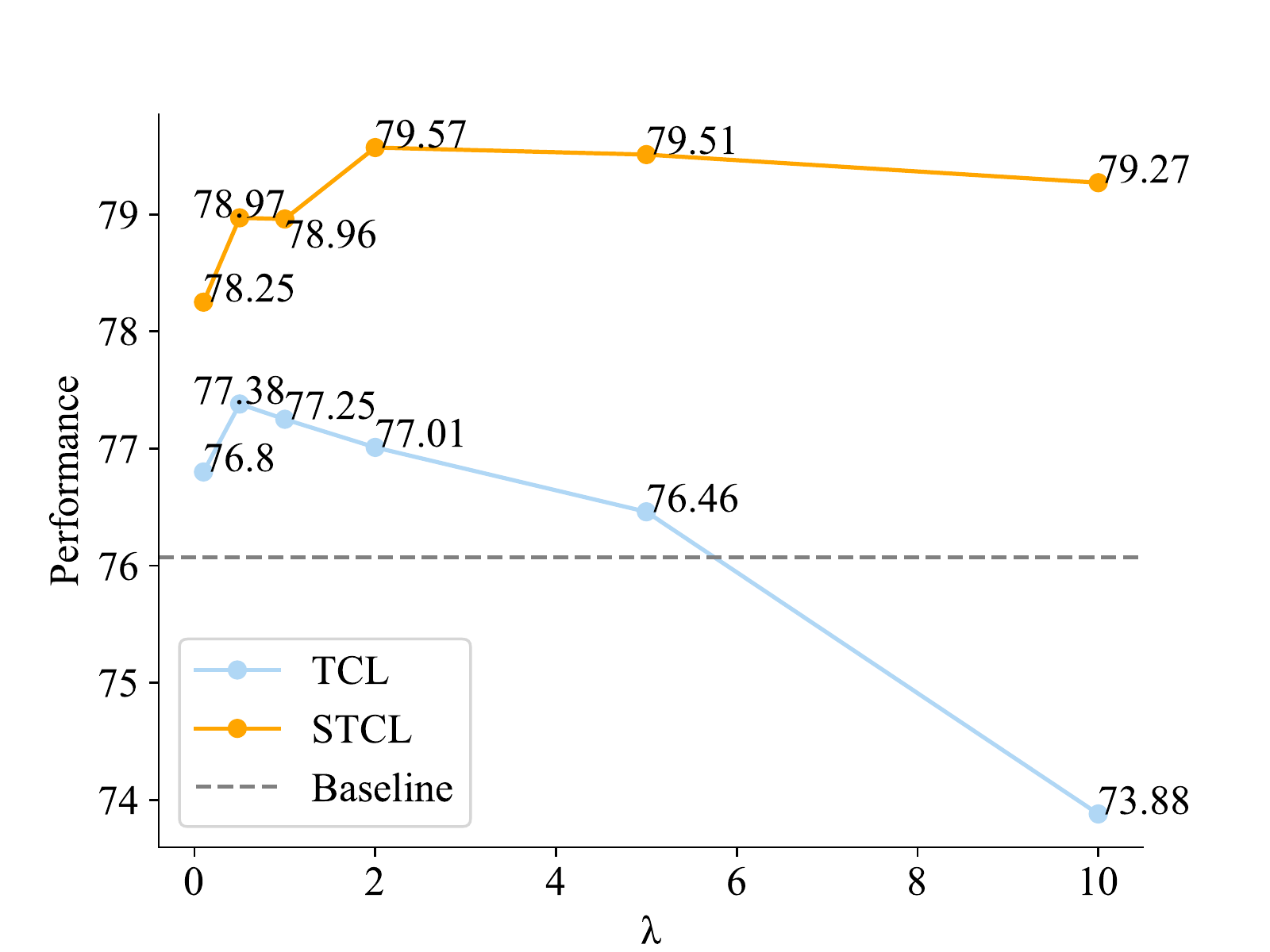}
  \end{subfigure}
  \caption{The effect of hyperparameters in contrastive loss on performance (\textbf{left}: $\tau$, \textbf{right}: $\lambda$).}
  \label{fig:combined}
  \vspace{-5mm}
\end{figure}

{\bf Influence of hyperparameter  $\tau$ and $\lambda$.}
The performance of our proposed frameworks are influenced by hyperparameters $\tau$ and $\lambda$ in the contrastive loss function. Fig. \ref{fig:combined} highlights the importance of tuning these hyperparameters while showcasing the robustness of our proposed frameworks to variations within a reasonable range (\emph{i.e.}, SEW-18). Fig. \ref{fig:combined} (left) shows the effect of varying $\tau$. As $\tau$ increases, both TCL and STCL display robust performance. Specifically, performance peaks at $\tau=0.07$ for TCL (77.38\%) and at $\tau=0.06$ for STCL (79.74\%). Fig. \ref{fig:combined} (right) examines the influence of the $\lambda$ parameter. For TCL, lower $\lambda$ values yield better results, with optimal performance at $\lambda=0.5$ (77.38\%). In contrast, STCL is more resilient to changes in $\lambda$, with the highest accuracy at $\lambda=2$ (79.57\%). This suggests that while the choice of $\lambda$ is critical, STCL's performance is less sensitive to this parameter, indicating good generalization ability. Based on these findings, we set $\tau$ to 0.07 and $\lambda$ to 0.5 for TCL, and 5 for STCL by default.

\begin{figure}[ht]
  \centering
  \begin{subfigure}{0.45\textwidth}
    \centering
    \includegraphics[scale=0.40]{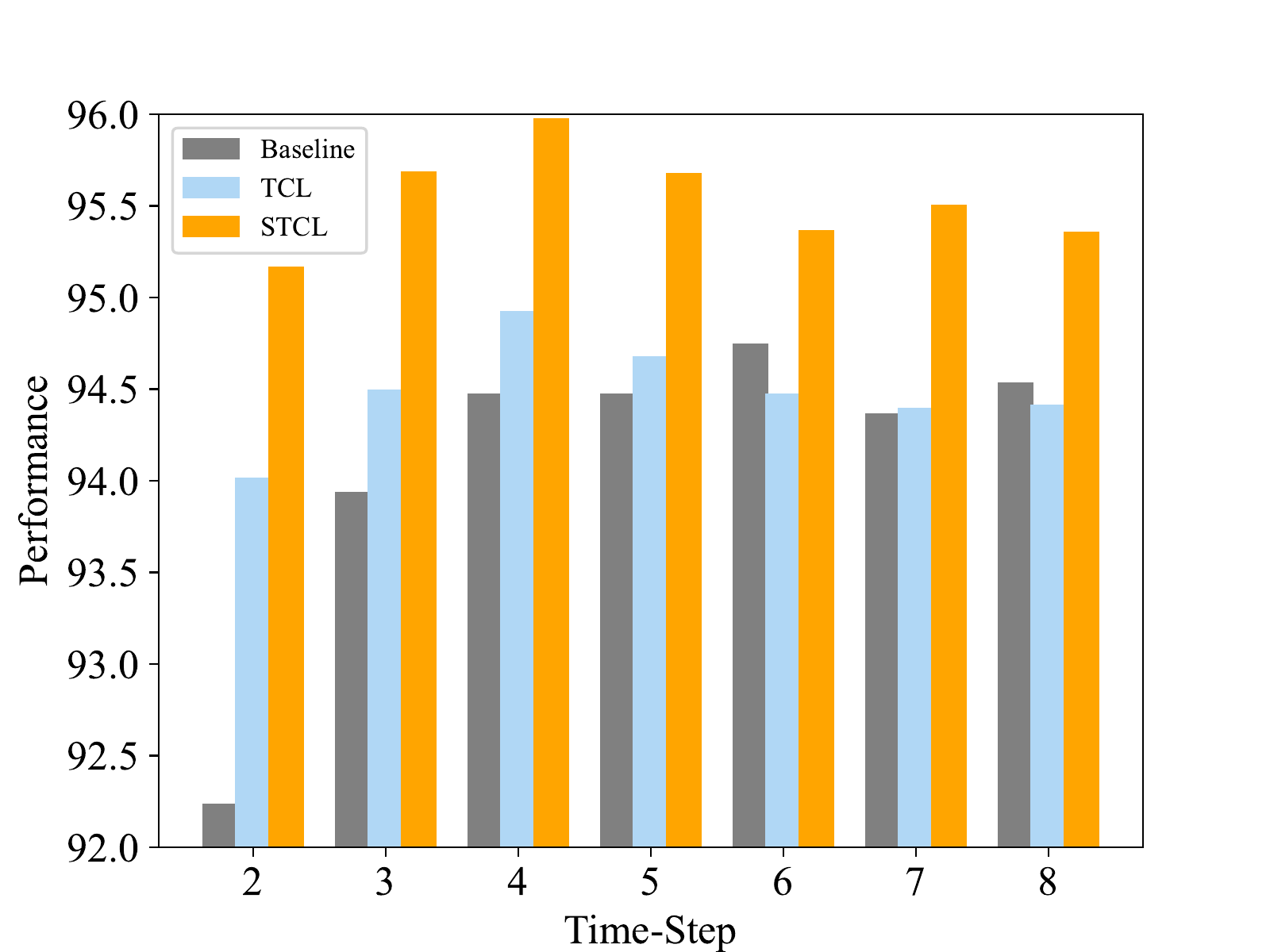}
  \end{subfigure}
   \hspace{0.02\textwidth}
  \begin{subfigure}{0.45\textwidth}
    \centering
    \includegraphics[scale=0.40]{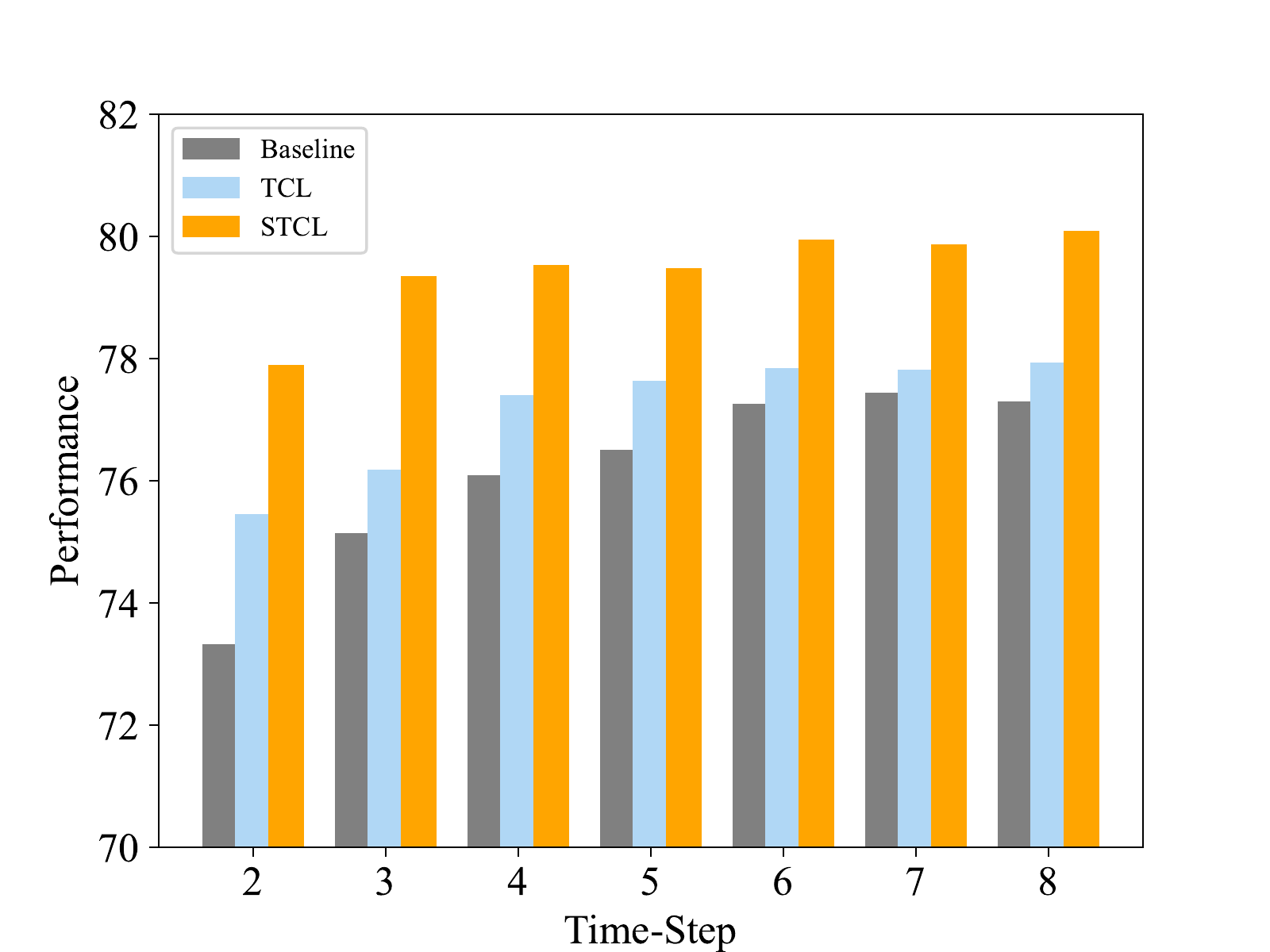}
  \end{subfigure}
  \caption{Influence of time step $T$ on performance (\textbf{left}: CIFAR-10, \textbf{right}: CIFAR-100).}
  \label{fig:combined_time}
  \vspace{-3mm}
\end{figure}

{\bf Influence of time steps $T$.} The effect of varying time steps ($T$) on the performance of our proposed methods, TCL and STCL, is examined on CIFAR-100 and CIFAR-10 datasets in this section. As depicted in Fig. \ref{fig:combined_time}, both TCL and STCL consistently exceed the baseline across different time steps for both datasets (\emph{i.e.}, SEW-18). This underlines the ability of our frameworks to enhance performance by effectively modeling the temporal correlations in the SNNs. Our STCL method, exploiting both data augmentation invariance and temporal correlation, achieves the highest accuracy on both datasets. Specifically, it reaches 80.07\% on CIFAR-100 and 95.97\% on CIFAR-10, with time steps of 8 and 4, respectively. However, the performance tends to plateau or slightly decrease with an increase in time steps beyond these optimal points, indicating a balance between the number of time steps and model performance. This suggests that while increasing time steps initially enhances the model's generalizability, excessive time steps could risk overfitting or computational inefficiencies.

{\bf Influence on spike firing rate.}
Tab. \ref{tab:fire_rate} dispalys the spike firing rate across SEW-18 blocks for our methods. TCL reveals a slight increase in fire rates, contributing to notable performance enhancements. Meanwhile, STCL shows varied trends across blocks, implying intricate feature extraction. Despite a slight rise in computation, the significant performance gain affirms the viability of our method.
\vspace{-5mm}
\begin{table}[ht]
\centering
\caption{Spike Firing rate across different SNN blocks}
\setlength\tabcolsep{1.9mm}
\renewcommand\arraystretch{1.1}
{
\scalebox{0.80}{
\begin{tabular}{cccccccccc}
\toprule
Block Index & 1 & 2 & 3 & 4 & 5 & 6 & 7 & 8 & Acc(\%)\\ 
\midrule
\rowcolor{gray!20} Baseline & 30.18 & 38.51 & 18.25 & 25.83 & 8.52 & 13.9 & 6.89 & 13.81 & 76.07\\
TCL & 32.98 & 43.98 & 22.23 & 32.83 & 11.52 & 18.84 & 10.33 & 22.68 & 77.38\\
STCL & 28.95 (\textcolor{green}{$\downarrow$})   & 34.28 (\textcolor{green}{$\downarrow$}) & 20.75 (\textcolor{red}{$\uparrow$}) & 28.11 (\textcolor{red}{$\uparrow$}) & 13.85 (\textcolor{red}{$\uparrow$}) & 20.46 (\textcolor{red}{$\uparrow$}) & 7.97 (\textcolor{red}{$\uparrow$}) & 22.20 (\textcolor{red}{$\uparrow$}) & 79.51 \\
\bottomrule
\end{tabular}
}}
\label{tab:fire_rate}
\vspace{-5mm}
\end{table}

\section{Conclusion}


To summarize, this study proposes innovative solutions to the inherent performance-latency trade-off problem in SNNs through the introduction of the TCL and STCL frameworks. It is demonstrated that these frameworks considerably enhance the ability of SNNs to make fully use of temporal correlations and learn invariance to data augmentation, thereby improving performance under extremely low latency conditions. Experimental results, drawn from eight datasets, validate the proposed methods, with performance significantly outperforming established SOTA approaches. Prospective research will concentrate on refining these frameworks and examining their applicability across different domains. Ultimately, the aim is to realize the vision of efficient, low energy computation of SNNs.

\clearpage

\end{document}